\documentclass[letterpaper]{article} % DO NOT CHANGE THIS
\usepackage{aaai2027}
\usepackage[hyphens]{url}  % DO NOT CHANGE THIS
\usepackage{graphicx} % DO NOT CHANGE THIS
\usepackage{natbib}  % DO NOT CHANGE THIS AND DO NOT ADD ANY OPTIONS TO IT
\usepackage{caption} % DO NOT CHANGE THIS AND DO NOT ADD ANY OPTIONS TO IT
\usepackage{algorithm}
\usepackage{algorithmic}

\usepackage{newfloat}
\usepackage{listings}
\DeclareCaptionStyle{ruled}{labelfont=normalfont,labelsep=colon,strut=off} % DO NOT CHANGE THIS
\floatstyle{ruled}
\newfloat{listing}{tb}{lst}{}
\floatname{listing}{Listing}

\usepackage{booktabs}

\usepackage{amssymb}
\usepackage{amsmath}
\usepackage{cuted}
\usepackage[table]{xcolor}
\definecolor{best}{HTML}{9ac4de}
\definecolor{second}{HTML}{d4e6f1}
\definecolor{urlblue}{HTML}{377EBD}
\usepackage{multirow}
\AtBeginDocument{
    \setlength{\abovedisplayskip}{4pt plus 2pt minus 2pt}
    \setlength{\belowdisplayskip}{4pt plus 2pt minus 2pt}
    \setlength{\abovedisplayshortskip}{5pt plus 2pt minus 2pt}
    \setlength{\belowdisplayshortskip}{5pt plus 2pt minus 2pt}
}

\title{RenderMatte: Exact-Alpha Rendering and Group-Relative Alignment for Image Matting}
\author{
    Zecheng Ren\textsuperscript{\rm 1},
    Yafei Hu\textsuperscript{\rm 2},
    Jianing Zhao\textsuperscript{\rm 1},
    Ruichen Cong\textsuperscript{\rm 1},
    Qun Jin\textsuperscript{\rm 1}\corresponding,
    Yiren Song\textsuperscript{\rm 3}
}
\affiliations{
    \textsuperscript{\rm 1}Waseda University\\
    \textsuperscript{\rm 2}Shanghai Jiao Tong University\\
    \textsuperscript{\rm 3}National University of Singapore\\
}

\begin{document}

\maketitle
\begingroup
\renewcommand{\thefootnote}{}
\footnotetext{*Corresponding Author.}
\endgroup

\begin{strip}
    \vspace{-14mm}
    \centering
    % 注意：删除了 Figures/ 前缀，因为图片就在根目录
    \includegraphics[width=\textwidth]{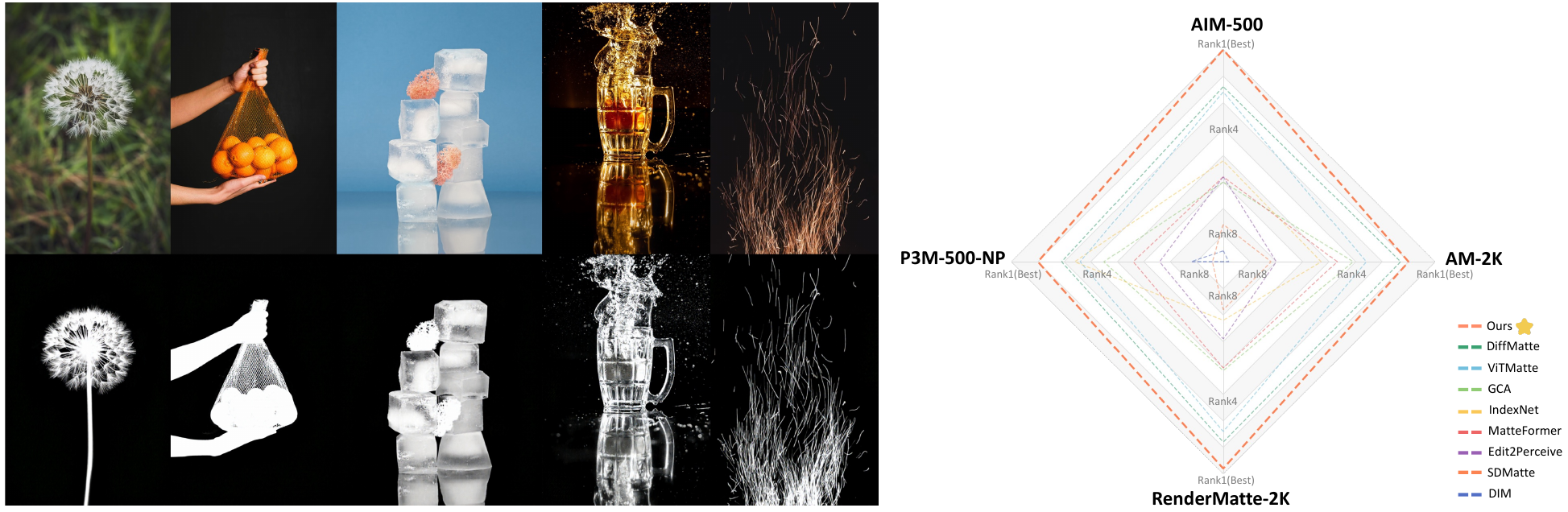} 
    \setlength{\abovecaptionskip}{-2mm}
    \captionof{figure}{We present \textbf{RenderMatte}, a trimap-guided image matting framework built upon image editing priors. Our model achieves state-of-the-art performance across the zero-shot matting task, consistently outperforming previous methods. Radar chart summarizing average rank performance across four benchmarks; points closer to the outer edge indicate superior performance. \textbf{Zoom in for better view.}}
    \label{fig:teaser}
\end{strip}

% \begin{figure*}[t] % 星号表示跨栏，通常跨栏图放在页面顶部 [t] 最美观
%     \centering
%     \includegraphics[width=\textwidth]{Figures/teaser_tight_balanced.pdf} % \textwidth 表示占据整页宽度
%     \caption{
%  We present \textbf{AlphaGRPO}, a reward-aligned image matting framework built upon image editing priors. Our model achieves state-of-the-art performance across the zero-shot matting task, consistently outperforming
% previous methods. Given diverse input images, our model predicts high-quality alpha mattes that preserve challenging sparse boundary structures, including hair, fur, and thin silhouettes.  The ranking pyramid on the right summarizes the average ranking score of our method against prior representative matting methods across AIM-500, P3M-500-NP, and AM-2K; lower is better.}
%     \label{fig:overview}
% \end{figure*}

\begin{abstract}
Image matting is an essential enabling technology for modern visual content production,
where foreground extraction determines the realism and editability of downstream
creation workflows. However, precise alpha estimation in open-world scenes
remains challenging because real foregrounds exhibit highly diverse appearances and opacity patterns. This makes existing methods struggle with semantic ambiguity and fine-grained opacity variation, especially in sparse boundary regions that are fragile and difficult to supervise. To address this gap, we present \textbf{RenderMatte}, a trimap-guided matting framework that adapts FLUX.1 Kontext through full-parameter fine-tuning, leveraging image editing priors for structure-preserving alpha prediction. During supervised adaptation, an alpha-edge objective preserves the latent flow-matching signal while strengthening pixel-space boundary supervision. We further introduce group-relative alpha alignment for post-training.
It compares multiple mattes sampled under the same trimap condition using
matting-specific rewards for alpha accuracy, boundary fidelity, trimap
compliance, and compositional consistency. To overcome the lack of precise edge annotations, we construct the \textbf{RenderMatte dataset}, a large-scale synthetic dataset combining 3D-rendered RGBA foregrounds with diverse multi-source assets. It features exact strand-level alpha annotations and diverse background composites. Experiments show state-of-the-art performance across all benchmarks,
demonstrating a scalable path toward high-fidelity matting in open-world scenes.
% Code is released at \textcolor{urlblue}{\url{https://github.com/Nislab-r/RenderMatte}}.

\end{abstract}

\vspace{0.6em}
\noindent\textbf{Code} --- \url{https://github.com/Nislab-r/RenderMatte}

\noindent\textbf{Dataset} ---\\
\hspace*{1.8em}\url{https://huggingface.co/datasets/Renz-7/RenderMatte-dataset}

\section{Introduction}

Image matting is an essential enabling technology for modern visual content
production. By estimating a continuous per-pixel alpha matte, it separates
foreground content from its background and supports realistic compositing,
interactive editing, virtual production, and generative creation workflows
~\cite{levin2008closedform,xu2017deepmatting,sun2023sparsemat,dai2025transadapter,yin2026qwenlayered}.
Unlike semantic segmentation, matting must recover fractional opacity at mixed
pixels, which makes the problem severely under-constrained. Practical systems
therefore often rely on auxiliary guidance, such as trimaps, to mark definite
foreground, definite background, and unknown regions where alpha estimation is
needed~\cite{xu2017deepmatting,park2023maskguided,ye2024unifying,huang2025sdmatte}.

However, trimap guidance does not remove the core difficulty of matting in
open-world scenes. Real foregrounds exhibit highly diverse appearances and
opacity patterns, while the most difficult regions are often visually small but
perceptually decisive. These regions require both semantic understanding and
precise low-level opacity estimation to avoid foreground-background confusion
and preserve boundary details. Existing matting pipelines still struggle with
this combination due to the scarcity of accurate high-frequency alpha
supervision~\cite{burgert2024magick,chen2026alpha}.

Large generative models provide a promising way to strengthen the missing
visual prior. They learn rich semantic and structural knowledge from large-scale
image data, which can help resolve ambiguous foreground appearances
~\cite{ramesh2022hierarchical,saharia2022photorealistic,rombach2022high,
podell2024sdxl,ke2024marigold,hu2024diffmatte}. Yet matting is not a
free-form generation task, since the output must stay spatially aligned with the
input and represent opacity rather than RGB appearance. This makes image editing
models especially relevant, as they are trained to transform an input image
while preserving unrelated structures, making them better suited to
structure-preserving alpha prediction~\cite{brooks2023instructpix2pix,zhang2023controlnet,edit2perceive}.

To address these challenges, we develop \textbf{RenderMatte}, a unified
framework that builds on the image-editing prior of FLUX.1 Kontext
~\cite{labs2025flux}. Given an RGB image and a trimap, we formulate matting as a
conditional image editing task and fine-tune the model to produce a pixel-aligned
alpha matte instead of an RGB image. This adaptation transfers semantic and
structural priors to fractional-opacity prediction. However, supervised fine-tuning may still overfit to synthetic compositing artifacts or background context rather than the actual foreground structures. To force the model to focus on foreground-intrinsic boundaries, we therefore introduce a matting-specific alpha-edge loss to enforce
alpha consistency across image recomposition and repeated matting, encouraging
the model to recover foreground-intrinsic opacity structures and fine details.

We further introduce group-relative alpha alignment for post-training, inspired
by recent policy optimization methods for visual generation
~\cite{black2023ddpo,wallace2024diffusiondpo,xue2025dancegrpo}.
Standard regression losses optimize a single prediction with absolute
pixel-wise errors, so they are often dominated by large easy regions with alpha
values near 0 or 1. Boundary-weighted losses reduce this imbalance, but still
treat each prediction independently. In contrast, our generative model produces
multiple candidate mattes for the same input. A task-specific reward compares
these candidates by alpha precision, boundary quality, trimap compliance, and
compositional consistency. This relative comparison encourages the model to favor predictions with better overall matting quality and finer boundary details.

Finally, we construct the \textbf{RenderMatte dataset} to address the data
bottleneck behind high-fidelity matting. Existing benchmarks are too small or
category-biased to capture real-world foreground diversity, while training data
often lack precise supervision around sparse boundaries
~\cite{aim500,p3m10k,am2k,burgert2024magick}. To overcome this, we build a
scalable synthesis pipeline centered on multi-source RGBA foregrounds. The main
source is 3D-rendered foreground assets, which provide exact strand-level alpha
annotations for fine structures such as hair and fur~\cite{greff2022kubric,roberts2021hypersim,deitke2022objaverse}. We also include other
RGBA foreground assets to broaden the foreground distribution. These foregrounds
are composited with diverse natural backgrounds, producing reliable data for both supervised adaptation and reward-based post-training.

Our main contributions are summarized as follows:

\begin{itemize}
    \item We introduce \textbf{RenderMatte}, a trimap-guided matting framework
    that adapts FLUX.1 Kontext through full-parameter fine-tuning, with an
    alpha-edge loss for foreground-intrinsic boundary preservation.

    \item We propose a group-relative alpha alignment strategy that treats sampled
mattes as competing candidates under the same trimap condition and optimizes
their relative quality using dense matting rewards.

     \item We construct the \textbf{RenderMatte dataset}, a multi-source synthetic
    matting dataset with $83{,}533$ training composites and a $2{,}000$-image
    test set, providing exact alpha supervision for transparent objects and
    complex boundaries.
\end{itemize}

\begin{figure*}[t] % 星号表示跨栏，通常跨栏图放在页面顶部 [t] 最美观
    \centering
    \includegraphics[width=\textwidth]{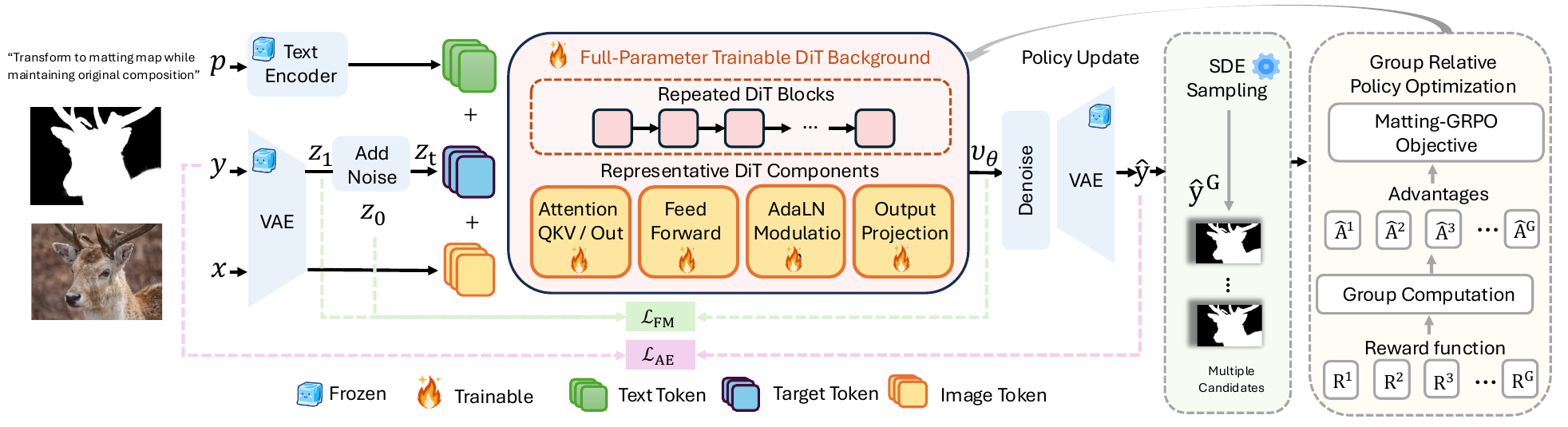} % \textwidth 表示占据整页宽度
    \setlength{\abovecaptionskip}{-2mm}
    \caption{
\textbf{Overview of the RenderMatte framework.} We adapt the FLUX.1 Kontext editor for
image matting~\cite{labs2025flux}. Given a text prompt \(p\), target image \(y\), and input image \(x\), the model predicts an alpha matte \(\hat{y}\). In the forward
process, the target alpha token is encoded as \(z_1\), noised into \(z_t\),
and concatenated with the text and image tokens. The trainable DiT
backbone predicts the velocity \(v_\theta\) from noise \(z_0\) to the target
latent \(z_1\). Supervised fine-tuning combines the latent-space flow matching
loss \(\mathcal{L}_{\mathrm{FM}}\) with the alpha-edge loss
\(\mathcal{L}_{\mathrm{AE}}\). We then perform group-relative alpha alignment: SDE sampling produces multiple
candidate mattes \(\{\hat{y}^{i}\}_{i=1}^{G}\), whose rewards
\(\{R^{i}\}_{i=1}^{G}\) are converted into relative advantages
\(\{\hat{A}^{i}\}_{i=1}^{G}\) through a GRPO-style objective.
}
    \label{fig:pipeline}
\end{figure*}

\begin{figure*}[t] % 星号表示跨栏，通常跨栏图放在页面顶部 [t] 最美观
    \centering
    \includegraphics[width=\textwidth]{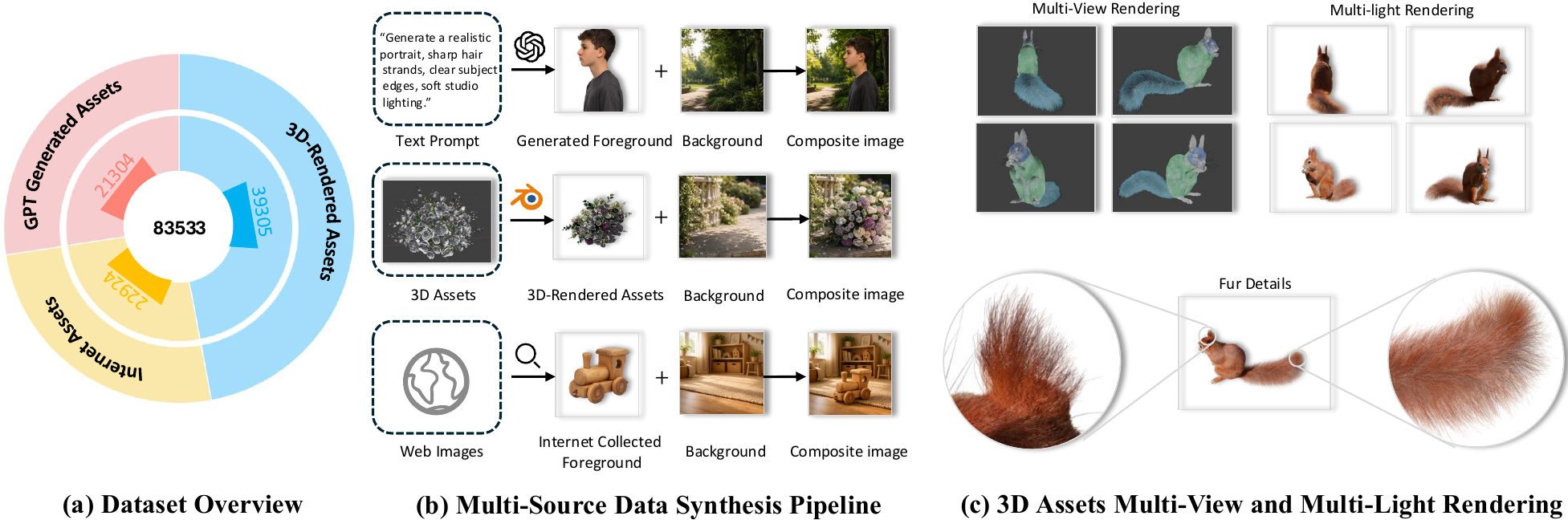} % \textwidth 表示占据整页宽度
    \setlength{\abovecaptionskip}{-2mm}
    \caption{
\textbf{Overview of RenderMatte dataset construction.}  (a) RenderMatte contains 83,533
composite training images from three foreground sources: 3D-rendered assets,
GPT-generated assets, and Internet-collected assets. (b) We synthesize
training samples by pairing each foreground with a background image. This pipeline preserves
exact alpha annotations while increasing foreground--background diversity.
(c) For 3D assets, we render each foreground under multiple camera viewpoints
and lighting directions, producing diverse silhouettes, shading patterns, and
strand-level hair or fur boundaries.
}
    \label{fig:dataset}
\end{figure*}

\section{Related Work}
\label{sec:related_work}

% \subsection{Image Matting}
% Classical matting methods estimate alpha mattes from low-level assumptions,
% such as color affinity and local sampling~\cite{levin2008closedform,chen2013knnmatting}.
% Deep matting methods replace these hand-crafted assumptions with learned
% alpha predictors trained on composited data~\cite{xu2017deepmatting}. Later
% methods improve the predictor with index-guided upsampling and contextual
% attention~\cite{lu2019indices,hou2019context,li2020natural}. Recent methods
% further introduce transformer backbones, pretrained ViTs, and stronger
% guidance mechanisms~\cite{park2022matteformer,yao2024vitmatte,ye2024unifying,park2023maskguided,guo2024incontext}.
% Despite different architectures and interaction forms, these methods remain
% constrained by the quality of alpha supervision. This is especially limiting for
% high-frequency boundaries, where manual labels are expensive and pseudo-labels
% often inherit smoothing and leakage artifacts. EdgeMatte addresses this
% data-side limitation with exact RGBA foregrounds and strand-level alpha
% annotations produced by 3D rendering.

\subsection{Image Matting}
Classical matting methods estimate alpha mattes from low-level assumptions,
such as color affinity and local sampling~\cite{levin2008closedform,chen2013knnmatting}.
Deep trimap-guided methods such as DIM, IndexNet, GCA, and FBA improve
low-level alpha prediction through learned regression, index-guided
upsampling, contextual attention, and foreground-background-alpha estimation
~\cite{xu2017deepmatting,lu2019indices,hou2019context, li2020natural,forte2020fba}. Trimap-free portrait matting further reduces user guidance through objective
decomposition~\cite{ke2022modnet}. These
methods directly target boundary reconstruction, but remain limited by the
scale and precision of alpha supervision. To strengthen semantic reasoning,
MatteFormer, ViTMatte, and unified matting frameworks introduce transformer
backbones, pretrained ViTs, and stronger guidance mechanisms
~\cite{chen2018semantic,lin2022robust,park2022matteformer,
yao2024vitmatte,guo2024incontext, ye2024unifying}. However,
high-frequency structures such as hair, fur, and thin objects remain
difficult when alpha supervision does not explicitly capture sparse boundary
details. RenderMatte addresses this with exact RGBA supervision, trimap-aware alpha-edge
loss, and group-relative alpha alignment.

\subsection{Image Generation and Editing Models}
To obtain stronger visual priors, recent dense prediction methods adapt large
generative models instead of relying only on task-specific predictors.
Diffusion and flow-based models learn broad visual knowledge from large-scale
image data~\cite{ho2020denoising,rombach2022high,lipman2022flow,podell2024sdxl}. These
models have been adapted to depth estimation and image matting, including
Marigold, DiffMatte, MattingGen, and SDMatte
~\cite{ke2024marigold,hu2024diffmatte,wang2024mattinggen,huang2025sdmatte}.
Many of these methods start from text-to-image generators, which are optimized
for concept-to-image synthesis rather than deterministic image-to-image
reasoning~\cite{edit2perceive}. Image editing diffusion models reduce this
mismatch because they condition on an existing image and learn
structure-preserving transformations~\cite{brooks2023instructpix2pix,zhang2023controlnet,labs2025flux}.
FE2E and Edit2Perceive validate this editing-based paradigm for dense
perception through image-to-image consistency and dense correspondence
~\cite{fe2e,edit2perceive}. This motivates our use of FLUX.1 Kontext for trimap-guided alpha prediction, where the output must remain spatially aligned with the input while representing opacity rather than RGB appearance~\cite{labs2025flux}.

\subsection{Reward-Based Post-Training for Visual Generation}
Post-training aligns generative models with objectives beyond likelihood
training. DDPO and DPOK formulate diffusion fine-tuning as policy optimization
with reward feedback~\cite{black2023ddpo,fan2023dpok}. Diffusion-DPO and
AlignProp instead optimize preference or differentiable rewards for
text-to-image generation~\cite{wallace2024diffusiondpo,prabhudesai2024alignprop}.
Recent GRPO-style methods further show that relative candidate quality can
provide a stable signal for visual generation alignment~\cite{liang2025spo, xue2025dancegrpo}.
These methods mainly optimize global generation quality, such as aesthetics,
text alignment, or preference scores. Image matting requires reference-based
dense quality instead: the output must be spatially aligned, numerically
precise, and sensitive to boundary errors. Different from prior GRPO-style visual generation methods that optimize global image preferences, we compare alpha mattes conditioned on the same image-trimap pair. The resulting relative advantage reflects dense alpha errors, boundary fidelity, connectivity, and trimap consistency.

\section{Method}
\label{sec:method}

This section presents the full RenderMatte pipeline. The key idea is to couple exact-alpha supervision with candidate-level reward alignment: supervised adaptation teaches the model to represent alpha mattes, while group-relative post-training selects among plausible sampled mattes using matting-specific dense quality signals.

\subsection{Overall Architecture}
\label{sec:method_overview}

RenderMatte adapts FLUX.1 Kontext from image editing to trimap-guided alpha
matting~\cite{labs2025flux}. Given an input image \(x \in \mathbb{R}^{H \times W \times 3}\),
trimap guidance \(g \in [0,1]^{H \times W}\), and a text prompt
\(p\), our goal is to predict the alpha matte \(y \in [0,1]^{H \times W}\). During training, \(g\) is derived from the alpha matte \(y\). We reshape the alpha matte target and trimap guidance into image-like inputs to match the image-editing backbone.

As shown in Fig.~\ref{fig:pipeline}, RenderMatte contains two training stages.
The first stage performs supervised adaptation, where latent flow matching
transfers the editing prior to alpha prediction and the alpha-edge loss adds
pixel-space boundary supervision. The second stage performs group-relative alpha alignment, where multiple
candidate mattes are sampled under the same image-trimap condition and ranked
by matting-specific rewards. This design separates alpha representation learning
from boundary-sensitive quality alignment.

The visual condition, formed by the input image and trimap guidance, is encoded
as \(c_{x,g}\), while the text instruction \(p\) is encoded as \(c_p\). During
supervised adaptation, the target dense map \(y\) is projected into a target
latent \(z_1\) via the frozen VAE encoder. To perform flow matching within the
DiT, a noisy latent \(z_t\) is constructed along a linear trajectory, patchified,
and encoded into target tokens. These target tokens, along with the image and
text tokens, are processed by the DiT backbone~\cite{10377858}.

\subsection{Alpha-Edge Supervised Objective}
\label{sec:alpha_edge_adaptation}

\paragraph{Flow-matching loss.}
We first adapt the generative editor to the matting domain using the rectified flow objective~\cite{liu2022flowstraightfastlearning}. We sample a pure Gaussian noise latent \(z_0 \sim \mathcal{N}(0,\mathbf{I})\) and construct a straight-line trajectory to define the intermediate noisy latent \(z_t\) at timestep \(t \in [0, 1]\):
\begin{equation}
    z_t = (1 - t) z_0 + t z_1.
    \label{eq:rectified_flow_path}
\end{equation}
Under this formulation, the constant velocity vector of this path is non-time-dependent and simplifies to \(\mathbf{v} = z_1 - z_0\). The DiT backbone \(\mathbf{v}_{\theta}\) is trained to predict this velocity, conditioned on the intermediate state \(z_t\), timestep \(t\), and encoded conditions \(c_{x,g}\) and \(c_p\), by minimizing the flow-matching objective:
\begin{equation}
\small
\mathcal{L}_{\mathrm{FM}}
=
\mathbb{E}_{t,z_1,z_0,c_{x,g},c_p}
\!\left[
\left\|
\mathbf{v}_{\theta}(\mathrm{concat}(z_t,c_{x,g},c_p),t)
-\mathbf{v}
\right\|_2^2
\right].
\label{eq:flow_matching}
\end{equation}
\vspace{-4mm}
\paragraph{Alpha-Edge loss.}
Latent flow matching provides stable generative adaptation, but it does not
directly enforce alpha accuracy in pixel space. We therefore decode the predicted latent into an
alpha matte \(\hat{y}\) and introduce an alpha-edge objective that explicitly separates uncertain and certain trimap regions while adding multi-scale boundary supervision on the ambiguous band.

Let \(\Omega_u\) and \(\Omega_k\) denote the uncertain and certain regions
defined by the trimap guidance \(g\). The alpha-edge loss is
\begin{equation}
\begin{aligned}
    \mathcal{L}_{\mathrm{AE}}
    =
    &\frac{1}{|\Omega_u|}
    \sum_{q \in \Omega_u}
    |\hat{y}_q-y_q|
    +
    \frac{1}{|\Omega_k|}
    \sum_{q \in \Omega_k}
    |\hat{y}_q-y_q| \\
    &+
    \lambda_{\mathrm{bd}}
    \mathcal{L}_{\mathrm{bd}}(\hat{y},y;\Omega_u),
\end{aligned}
\label{eq:alpha_edge_loss}
\end{equation}
where \(q\) indexes pixels, \(|\Omega|\) is the region size, and
\(\lambda_{\mathrm{bd}}\) weights the boundary-sensitive detail term
\(\mathcal{L}_{\mathrm{bd}}\), implemented as a multi-scale Laplacian pyramid loss over the uncertain region~\cite{burt1987laplacian}. This term encourages the model to recover high-frequency alpha structures across multiple spatial scales.

\paragraph{Adaptive loss weighting.}
We jointly optimize the latent-space and pixel-space objectives:
\begin{equation}
    \mathcal{L}_{\mathrm{SFT}}
    =
    \mathcal{L}_{\mathrm{FM}}
    +
    \lambda_{\mathrm{AE}}(s)
    \mathcal{L}_{\mathrm{AE}},
    \label{eq:sft_objective}
\end{equation}
where \(s\) denotes the training step. Instead of using a fixed weight, we
adaptively balance the two losses by their detached magnitudes:
\begin{equation}
    \lambda_{\mathrm{AE}}(s)
    =
    \frac{
        \mathrm{sg}(\mathcal{L}_{\mathrm{FM}})
    }{
        \mathrm{sg}(\mathcal{L}_{\mathrm{AE}})+\epsilon_{\mathrm{num}}
    }
    \gamma(s),
    \label{eq:adaptive_ae_weight}
\end{equation}
where \(\mathrm{sg}(\cdot)\) denotes stop-gradient and
\(\epsilon_{\mathrm{num}}\) is set to \(10^{-3}\) for numerical stability.
The schedule term is implemented as
\(\gamma(s)=\max(0, s/N_{\mathrm{iter}}-e_0)\), where \(N_{\mathrm{iter}}\)
is the number of iterations per epoch and \(e_0\) is the epoch at which the
pixel-space objective is enabled.
\subsection{Group-Relative Alpha Alignment}
\label{sec:reward_alignment}

\paragraph{GRPO-style policy objective.}
Although supervised adaptation provides a strong initialization, its fixed surrogate losses do not directly optimize the non-differentiable metrics used to evaluate matting quality. We therefore view the denoising trajectory that generates an alpha matte as a
policy, and use the relative quality among mattes sampled from the same
image-trimap condition as the optimization signal. For each conditioning context \(x_c=(c_{x,g},c_p)\), the current policy
\(\pi_{\theta}\) samples a group of \(G\) candidate alpha mattes
\(\{\hat{y}^{i}\}_{i=1}^{G}\).

Each candidate $\hat{y}^{i}$ is evaluated by a comprehensive matting reward function to receive a scalar score $R^i$. Instead of introducing a separate critic network, GRPO evaluates the trajectory-level relative advantage $\hat{A}^{i}$ by standardizing the rewards within the sampled group:
\begin{equation}
    \hat{A}^{i}=\frac{R^{i}-\mathrm{mean}(\{R^{j}\}_{j=1}^{G})
    }{\mathrm{std}(\{R^{j}\}_{j=1}^{G})+\epsilon_{\mathrm{std}}},
    \label{eq:grpo_advantage}
\end{equation}
where \(\epsilon_{\mathrm{std}}\) is a small constant for numerical stability.

% This trajectory-level advantage is then broadcast to all latent transitions along the corresponding sampling path. We optimize the policy parameters over the generative flow trajectory via a clipped surrogate objective with a Kullback-Leibler (KL) divergence penalty to prevent reward hacking:
% \begin{equation}
% \begin{aligned}
%     \mathcal{L}_{\mathrm{GRPO}}
%     =
%     -\mathbb{E}_{x_c, i, t}
%     \Big[&
%     \min\big(
%         \rho_{i,t}\hat{A}^{i},
%         \mathrm{clip}(\rho_{i,t},1-\epsilon,1+\epsilon)\hat{A}^{i}
%     \big) \\
%     &- \beta \mathbb{D}_{\mathrm{KL}}(\pi_{\theta} \| \pi_{\mathrm{ref}})
%     \Big],
% \end{aligned}
% \label{eq:grpo_objective}
% \end{equation}
% where $\beta$ controls the KL penalty strength against the reference model $\pi_{\mathrm{ref}}$ (initialized from the supervised adaptation stage), and the importance sampling ratio $\rho_{i,t}$ is defined as:
This trajectory-level advantage is then broadcast to all latent transitions along the corresponding sampling path. We optimize the policy parameters over the generative flow trajectory via a clipped surrogate objective:
\begin{equation}
\begin{aligned}
    \mathcal{L}_{\mathrm{GRPO}}
    =
    -\mathbb{E}_{x_c, i, t}
    \Big[
    \min\big(
        \rho_{i,t}\hat{A}^{i},
        \mathrm{clip}(\rho_{i,t},1-\epsilon,1+\epsilon)\hat{A}^{i}
    \big)
    \Big],
\end{aligned}
\label{eq:grpo_objective}
\end{equation}
where \(\epsilon\) is the clipping threshold and the importance sampling ratio $\rho_{i,t}$ is defined as:
\begin{equation}
    \rho_{i,t}
    =
    \exp
    \left(
        \log \pi_{\theta}(a_{i,t}\mid z_{i,t},x_c)
        -
        \log \pi_{\mathrm{old}}(a_{i,t}\mid z_{i,t},x_c)
    \right).
    \label{eq:grpo_ratio}
\end{equation}
Here, \(\pi_{\mathrm{old}}\) denotes the policy used to generate the sampled
trajectories before the current update. \(z_{i,t}\) represents the intermediate latent state at trajectory step
\(t\), and \(a_{i,t}\) denotes the sampled latent transition from
\(z_{i,t}\) to \(z_{i,t-1}\). Following generative policy optimization
paradigms~\cite{black2023ddpo,xue2025dancegrpo}, the network-predicted
velocity parameterizes the Gaussian transition mean used to compute the
log-probability of \(a_{i,t}\). The single image-level reward \(R^i\) is
uniformly shared across all trajectory steps for the same candidate.

\paragraph{Matting reward.}
We define the alpha-space reward to be consistent with the alpha-edge objective. Specifically, the
terminal reward uses the same family of alpha-error criteria, but converts the resulting errors into a bounded score for group-relative alpha comparison. For the \(i\)-th sampled candidate, we define:
\begingroup
\setlength{\abovedisplayskip}{7pt}
\setlength{\belowdisplayskip}{7pt}
\setlength{\abovedisplayshortskip}{2pt}
\setlength{\belowdisplayshortskip}{2pt}
\begin{equation}
    \vspace{-1mm}
    R^{i}
    =
    -
    \sum_{m \in \mathcal{M}_{\mathrm{R}}}
    w_m
    \min
    \left(
        \frac{
            e_m(\hat{y}^{i},y,g)
        }{
            s_m
        },
        \tau
    \right),
    \label{eq:matting_reward}
    \vspace{-1mm}
\end{equation}
where \(e_m(\cdot)\) is the error function for the \(m\)-th criterion in
\(\mathcal{M}_{\mathrm{R}}\), covering alpha accuracy, boundary fidelity,
structural consistency, and trimap compliance. The negative sign converts lower errors into higher rewards; \(w_m\), \(s_m\), and \(\tau\) denote the criterion weight, normalization scale, and clipping threshold, respectively.

Unlike generic image-level aesthetic or preference rewards, this reward is
defined entirely in alpha space and explicitly depends on the trimap partition,
thereby penalizing errors in the unknown boundary and violations in known
foreground/background regions.

\subsection{Large-Scale Synthetic Matting Data}
\label{sec:data_synthesis}

\paragraph{Multi-source foreground construction.}
High-quality alpha supervision is difficult to obtain at scale, especially
for hair, fur, thin objects, and transparent or semi-transparent regions.
Existing matting datasets provide limited coverage of these challenging alpha
structures, despite their importance for high-fidelity matting. Following the
common compositing-based data construction paradigm in image
matting~\cite{xu2017deepmatting,am2k}, we construct the RenderMatte dataset
through a multi-source data synthesis pipeline, as shown in Fig.~\ref{fig:dataset}.
The foreground asset pool contains \(25{,}211\) RGBA foregrounds from three
sources: 3D-rendered assets, GPT-generated foreground assets, and
Internet-collected foregrounds. During asset construction, we deliberately
increase the proportion of transparent and semi-transparent foregrounds, which
introduce complex alpha transitions and boundary ambiguities. Each foreground is
composited with sampled background images using its alpha matte to produce
training samples.

We synthesize more composites from 3D-rendered assets because they provide
exact high-frequency alpha annotations and controllable rendering conditions~\cite{greff2022kubric,roberts2021hypersim,
deitke2022objaverse}.
For each 3D asset, we render foregrounds under different camera viewpoints and
lighting directions, increasing the diversity of poses, silhouettes, shading
patterns, and boundary appearances. This process preserves fine hair and fur
geometry with strand-level alpha details. Each 3D foreground
produces about eight composites on average, while other assets produce two to
four composites. After filtering failed synthesis cases, RenderMatte contains
\(83{,}533\) training composites.

\paragraph{Alpha compositing and guidance generation.}
Given a foreground color \(F\), alpha matte \(y\), and background image
\(B\), we synthesize the composite image \(I\) by alpha compositing~\cite{porter1984compositing}. Background images are drawn from the source adopted by BG-20K~\cite{am2k}. This procedure increases
foreground--background diversity while preserving exact alpha annotations.

During training, we generate trimap guidance from the ground-truth
alpha matte. Specifically, we apply grayscale morphological erosion and
dilation with a randomly sampled radius to obtain definite foreground,
definite background, and unknown regions. The resulting trimap is encoded as
\(1\), \(0\), and \(0.5\) for foreground, background, and unknown pixels,
respectively. Randomizing the morphology radius exposes the model to unknown
regions with different widths.

\section{Experiments}
\label{sec:experiments}

\begin{table*}[t]
    \centering
    \caption{
    Quantitative comparison on AIM-500, P3M-500-NP, AM-2K, and RenderMatte-2K.
    Lower values are better for all metrics.
    AvgRank is computed over the reported metrics with tied ranks averaged.
    The \colorbox{best}{best} and \colorbox{second}{second-best} results are highlighted. Methods marked with \(^{\dagger}\) use their native interaction format.
    }
    \label{tab:main_results_1024}
    \resizebox{\textwidth}{!}{
        \begin{tabular}{l|ccccc|ccccc|ccccc|ccccc|c}
        \hline
        \multirow{2}{*}{\centering\textbf{Method}}
        & \multicolumn{5}{c|}{\textbf{AIM-500}}
        & \multicolumn{5}{c|}{\textbf{P3M-500-NP}}
        & \multicolumn{5}{c|}{\textbf{AM-2K}}
        & \multicolumn{5}{c|}{\textbf{RenderMatte-2K}}
        & \multirow{2}{*}{\centering\textbf{AvgRank$\downarrow$}} \\
        & \textbf{MSE$\downarrow$} & \textbf{MAD$\downarrow$}
        & \textbf{SAD$\downarrow$} & \textbf{Grad$\downarrow$}
        & \textbf{Conn$\downarrow$}
        & \textbf{MSE$\downarrow$} & \textbf{MAD$\downarrow$}
        & \textbf{SAD$\downarrow$} & \textbf{Grad$\downarrow$}
        & \textbf{Conn$\downarrow$}
        & \textbf{MSE$\downarrow$} & \textbf{MAD$\downarrow$}
        & \textbf{SAD$\downarrow$} & \textbf{Grad$\downarrow$}
        & \textbf{Conn$\downarrow$}
        & \textbf{MSE$\downarrow$} & \textbf{MAD$\downarrow$}
        & \textbf{SAD$\downarrow$} & \textbf{Grad$\downarrow$}
        & \textbf{Conn$\downarrow$}
        & \\
        \hline

        DIM
        & 0.017 & 0.031 & 53.1 & 42.3 & 37.9
        & 0.007 & 0.014 & 24.5 & 29.2 & 20.6
        & 0.013 & 0.024 & 40.5 & 32.7 & 33.3
        & 0.019 & 0.032 & 34.0 & 39.1 & 22.1
        & 8.6 \\

        GCA
        & 0.021 & 0.051 & 23.4 & 15.9 & 12.1
        & 0.012 & 0.038 & \cellcolor{best}6.2 & 9.8 & \cellcolor{best}4.5
        & 0.007 & 0.031 & \cellcolor{best}6.3 & \cellcolor{second}5.5 & \cellcolor{best}4.8
        & 0.004 & 0.008 & 8.4 & 12.1 & 5.0
        & 4.9 \\

        IndexNet
        & 0.007 & 0.015 & 25.8 & 16.3 & 17.1
        & 0.001 & 0.004 & 7.2 & \cellcolor{second}9.5 & 5.5
        & 0.002 & 0.005 & 9.2 & 8.1 & 7.3
        & 0.005 & 0.011 & 11.4 & 16.0 & 7.2
        & 5.2 \\

        MatteFormer
        & 0.009 & 0.016 & 27.0 & 20.3 & 15.1
        & 0.002 & 0.005 & 7.9 & 11.8 & 5.7
        & 0.002 & 0.005 & 8.5 & 7.2 & 6.3
        & 0.004 & 0.008 & 8.6 & 13.9 & 4.7
        & 5.3 \\

        ViTMatte
        & \cellcolor{second}0.004 & 0.011 & 17.8 & 13.7 & \cellcolor{second}11.0
        & 0.001 & 0.004 & 7.1 & 10.6 & 5.0
        & 0.001 & 0.005 & 7.7 & 6.2 & 5.5
        & 0.001 & 0.004 & 4.5 & 4.8 & \cellcolor{second}3.0
        & 3.1 \\

        DiffMatte
        & 0.004 & \cellcolor{second}0.011 & \cellcolor{second}17.2 & \cellcolor{second}13.5 & 11.6
        & \cellcolor{second}0.001 & \cellcolor{second}0.004 & \cellcolor{second}6.6 & 10.0 & \cellcolor{second}4.8
        & \cellcolor{second}0.001 & \cellcolor{second}0.004 & \cellcolor{second}6.8 & \cellcolor{second}5.7 & \cellcolor{best}5.0
        & \cellcolor{second}0.001 & \cellcolor{second}0.004 & \cellcolor{second}4.3 & \cellcolor{second}4.4 & 3.0
        & \cellcolor{second}2.5 \\

        \hline

        SDMatte\(^{\dagger}\)
        & 0.011 & 0.019 & 31.8 & 26.8 & 17.5
        & 0.013 & 0.018 & 32.0 & 20.4 & 20.8
        & 0.006 & 0.010 & 17.5 & 13.2 & 10.9
        & 0.008 & 0.012 & 12.1 & 20.4 & 3.6
        & 7.7 \\

        Edit2Perceive\(^{\dagger}\)
        & 0.006 & 0.017 & 29.1 & 18.2 & 15.7
        & 0.003 & 0.011 & 19.4 & 13.2 & 10.2
        & 0.004 & 0.012 & 20.4 & 9.6 & 9.9
        & 0.005 & 0.011 & 11.5 & 9.2 & 5.0
        & 6.4 \\

        \hline

        RenderMatte
        & \cellcolor{best}0.002 & \cellcolor{best}0.008 & \cellcolor{best}13.6 & \cellcolor{best}11.2 & \cellcolor{best}9.8
        & \cellcolor{best}0.001 & \cellcolor{best}0.004 & \cellcolor{best}6.2 & \cellcolor{best}8.2 & \cellcolor{best}4.8
        & \cellcolor{best}0.001 & \cellcolor{best}0.004 & \cellcolor{best}6.5 & \cellcolor{best}5.5 & \cellcolor{second}5.2
        & \cellcolor{best}0.001 & \cellcolor{best}0.003 & \cellcolor{best}3.6 & \cellcolor{best}3.0 & \cellcolor{best}2.6
        & \cellcolor{best}1.6 \\

        \hline
    \end{tabular}
    }
\end{table*}

\begin{figure*}[t]
    \centering
    \includegraphics[width=\textwidth]{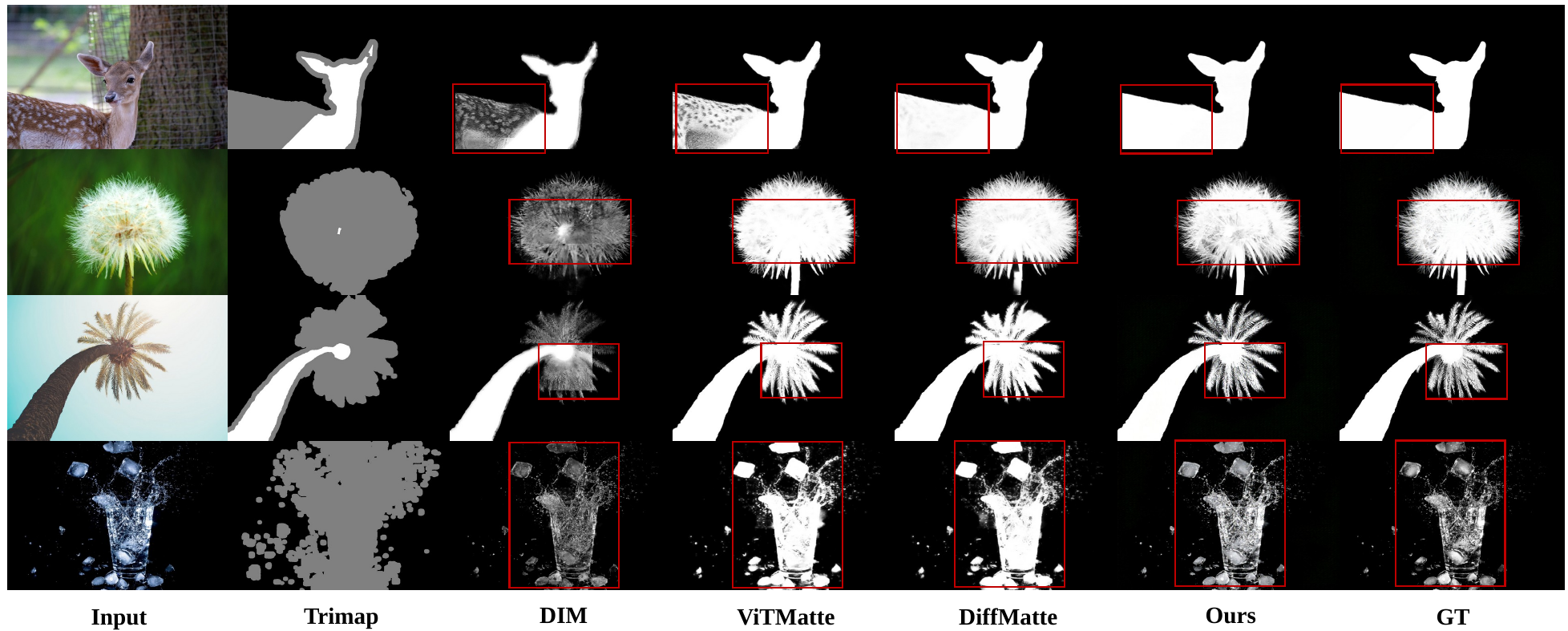}
    \setlength{\abovecaptionskip}{-2mm}
    \caption{
    Qualitative comparison with state-of-the-art image matting methods.
    From left to right, we show the input image, trimap, predictions from
    competing methods, our result, and the ground-truth alpha matte. \textbf{Zoom in for better view.}
    }
    \label{fig:qualitative_comparison}
\end{figure*}

\paragraph{Implementation details.}
Our model is built upon FLUX.1 Kontext~\cite{labs2025flux}. During supervised fine-tuning, we fully fine-tune the DiT backbone and keep the text encoders and VAE frozen. Images, trimaps, and alpha mattes are resized to \(1024\times1024\). We optimize the model with 8-bit AdamW using a learning rate of \(1\times10^{-5}\), weight decay \(0.01\), bf16 mixed precision, and DeepSpeed ZeRO-2 on 8 NVIDIA H200 GPUs. The per-device batch size is 3, giving an effective batch size of 24. The training objective combines the rectified-flow matching loss, restricted to the trimap unknown region, with the proposed alpha-edge loss.

The group-relative alignment stage is initialized from the supervised fine-tuned checkpoint. We freeze the base model and train a rank-64 LoRA adapter on the DiT modules. For each input condition, we sample \(G=8\) candidate alpha mattes using 8 denoising steps, guidance scale \(1.0\), and noise level \(0.2\). We instantiate the alignment stage with a GRPO-style clipped policy objective, using a clipping range of \(1\times10^{-4}\) and no KL penalty. The alignment adapter is optimized with 8-bit AdamW in bf16, using one inner epoch per sampling epoch. Unless otherwise specified, we use a learning rate of \(2\times10^{-4}\), an effective training batch size of 64, and evaluate checkpoints with one denoising step and guidance scale \(1.0\), without test-time ensembling.

\paragraph{Datasets and benchmarks.}
We train RenderMatte on our synthesized dataset. For evaluation, we use
AIM-500~\cite{aim500}, P3M-500-NP~\cite{p3m10k}, and AM-2K~\cite{am2k}. We also evaluate on RenderMatte-2K, a $2{,}000$-sample benchmark using completely unseen foregrounds and backgrounds to test zero-shot generalization.

\begin{figure}[t]
    \centering
    \includegraphics[width=0.85\linewidth,height=0.24\textheight,keepaspectratio]
    {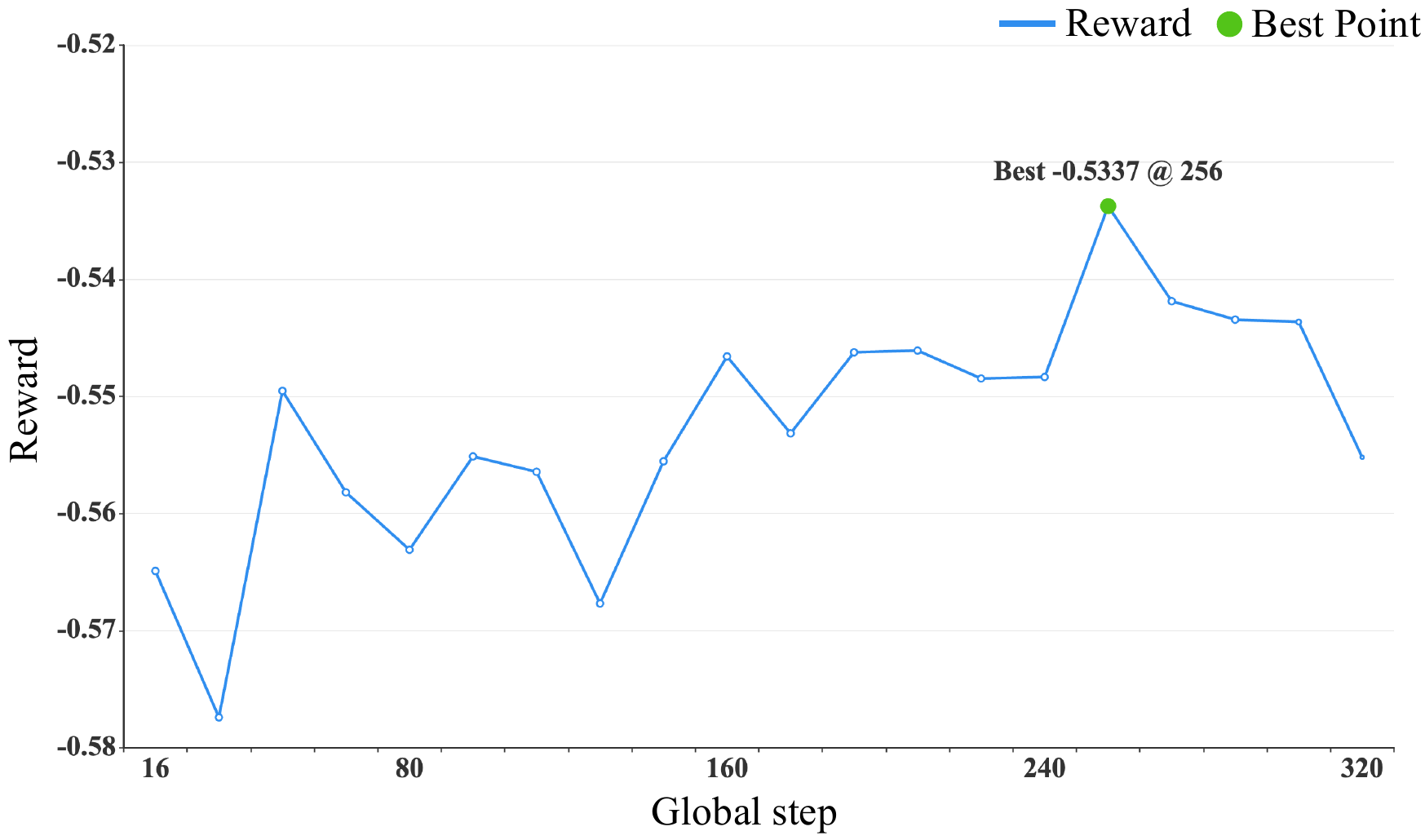}    
    \vspace{-2mm}
    \caption{Evaluation reward curve of group-relative alpha alignment on AIM-500.}
    \label{fig:reward_curve}
    \vspace{-4mm} % 控制：标题 与 下方正文 的间距 (数值请根据实际效果微调)
\end{figure}

\begin{figure*}[t]
    \centering
    \includegraphics[width=\textwidth]{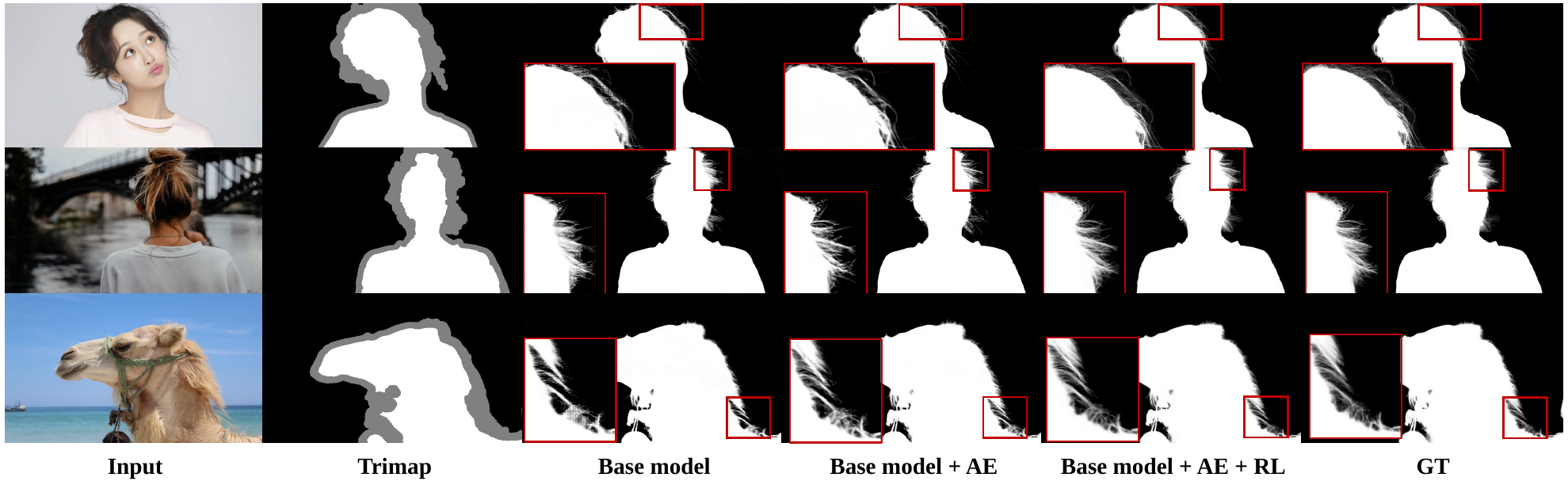}
    \setlength{\abovecaptionskip}{-2mm}
    \caption{
Visual ablation of the proposed training stages. From left to right, we show
the input image, trimap, predictions from different training stages, and the
ground-truth alpha matte. \textbf{Zoom in for better view.}
}
    \label{fig:ablation_comparison}
\end{figure*}

\paragraph{Evaluation metrics.}
We adopt five standard image matting metrics: mean squared error
(\textbf{MSE}), mean absolute difference (\textbf{MAD}), sum of absolute
differences (\textbf{SAD}), gradient error (\textbf{Grad}), and
connectivity error.

\subsection{Quantitative and Qualitative Evaluation}
\label{sec:main_results}

\paragraph{Quantitative comparison.}

Table~\ref{tab:main_results_1024} compares different methods across four benchmarks. Despite the challenges of adapting generative models for precise fractional-opacity regression, RenderMatte achieves the best average rank of 1.6 and obtains the lowest errors on most evaluated metrics. The consistent reduction in Grad and SAD indicates that exact-alpha synthetic supervision, alpha-edge pixel-space refinement, and group-relative alpha alignment improve both global alpha accuracy and boundary-sensitive detail recovery. Its strong performance on public benchmarks further suggests robust generalization to diverse foreground categories and boundary structures.

We also use the reward trajectory to examine why reward-based post-training is
useful beyond supervised adaptation. After supervised adaptation has saturated
under differentiable losses, the reward curve in Fig.~\ref{fig:reward_curve}
continues to improve and reaches its best value at step 256. This indicates
that reward-driven exploration can improve alpha predictions that are not
optimized by supervised training, supporting group-relative alignment as a post-training step rather than an auxiliary supervised loss.

\paragraph{Qualitative comparison.}
Figure \ref{fig:qualitative_comparison} compares the alpha mattes predicted by
RenderMatte and state-of-the-art baselines. As shown in the
challenging examples, existing methods often struggle with sparse or
semi-transparent structures, producing coarse silhouettes, incomplete object
interiors, oversmoothed boundaries, or missing low-opacity details.

In contrast, RenderMatte preserves finer alpha structures across diverse
foregrounds. It better retains animal contours, dandelion filaments, palm leaf
boundaries, and transparent water splashes, while maintaining cleaner definite
foreground regions. These visual results are consistent with our quantitative
boundary-sensitive improvements, confirming that editing priors, alpha-edge
supervision, and reward-aligned optimization jointly recover sparse boundary
details.

\subsection{Ablation Studies}
\label{sec:ablation}

To separate the effect of model adaptation, pixel-space alpha supervision, and
reward alignment, we evaluate the same FLUX.1 Kontext backbone under three
training configurations. We first examine whether adding
pixel-space supervision to latent flow matching improves sparse-boundary
recovery, and then evaluate whether reward-aligned post-training provides
additional gains beyond supervised adaptation. Figure \ref{fig:ablation_comparison} visualizes these progressive improvements.

% The results show that the two stages play
% different roles: alpha-edge supervision mainly improves boundary formation,
% while group-relative alpha alignment further refines candidate mattes that are already structurally
% plausible.

\paragraph{Effect of Alpha-Edge Supervision.}
As shown in Table~\ref{tab:ablation_training}, adding alpha-edge supervision
brings the main improvement over the baseline and improves all matting metrics.
The clearest gain appears on Grad, which directly reflects local alpha
transitions and boundary sharpness. This indicates that latent flow matching
alone can learn an alpha prior, but tends to smooth sparse high-frequency
regions. By adding pixel-space edge supervision, the model recovers sharper
contours and more stable boundaries. In Fig.~\ref{fig:ablation_comparison},
this effect is visible in the red boxes, where the base model misses or blurs
thin boundary structures, while the alpha-edge model restores clearer
silhouettes and fine strands.

\paragraph{Effect of Group-Relative Alpha Alignment.}
Group-relative alpha alignment further improves the alpha-edge model, mainly reducing SAD and Grad while also bringing a smaller gain on Conn. This suggests that reward alignment
acts as a refinement stage rather than a coarse correction stage. After
supervised adaptation has produced plausible alpha mattes, candidate-level
reward ranking encourages the model to prefer outputs with better structural
connectivity, sharper local transitions, and fewer boundary artifacts. As shown
in Fig.~\ref{fig:ablation_comparison}, this alignment stage does not drastically change the
overall foreground shape. Instead, it strengthens already recovered details in
the red-box regions, especially thin hair-like structures, low-opacity edges,
and fragmented boundary components. This qualitative difference explains why
this reward-based refinement complements alpha-edge supervision instead of simply duplicating its
effect.

\subsection{Failure Cases and Limitations}
\label{sec:limitations}

Despite the scale of our dataset, real-world foregrounds are virtually unlimited. RenderMatte may thus struggle with rare out-of-domain objects. Inference efficiency is another limitation: processing a \(1024 \times 1024\) image takes \(\sim\)\(2.7\)s on an A100 GPU, compared to \(<\)\(0.5\)s for DiffMatte. Efficient boundary-preserving inference remains future work.

% Despite the scale and diversity of our training dataset, real-world foregrounds remain virtually unlimited. RenderMatte may therefore struggle with rare or extreme out-of-domain objects, as shown in the supplementary failure cases. Inference efficiency is another limitation. Processing a \(1024 \times 1024\) image takes about \(2.7\) seconds on a single NVIDIA A100 GPU, compared with less than \(0.5\) seconds for DiffMatte. Improving efficiency while preserving boundary quality remains future work.

\begin{table}[t]
    \centering
    \caption{
    Ablation study averaged across all benchmarks.
    Lower values are better for all metrics.
        The \colorbox{best}{best} results are highlighted.
    }
    \label{tab:ablation_training}
    \resizebox{\columnwidth}{!}{
    \begin{tabular}{lcc|ccccc}
        \hline
        Base model & Alpha-Edge & RL
        & MSE & MAD & SAD & Grad & Conn \\
        \hline
        FLUX.1 Kontext
        & 
        & 
        & 0.0030 & 0.0073 & 11.6 & 14.5 & 7.5 \\

        FLUX.1 Kontext
        & \checkmark
        & 
        & 0.0023 & 0.0063 & 10.1 & 11.6 & 7.0 \\

        FLUX.1 Kontext
        & \checkmark
        & \checkmark
        & \cellcolor{best}0.0022
        & \cellcolor{best}0.0061
        & \cellcolor{best}9.6
        & \cellcolor{best}10.7
        & \cellcolor{best}6.7 \\
        \hline
    \end{tabular}
    }
\end{table}

\section{Conclusion}
\label{sec:conclusion}

We present RenderMatte, a trimap-guided matting framework that adapts FLUX.1 Kontext for structure-preserving alpha prediction. By combining trimap-aware alpha-edge supervision with group-relative alpha alignment, RenderMatte links exact-alpha data construction, supervised adaptation, and reward-based post-training in a unified matting pipeline. The RenderMatte dataset provides large-scale multi-source synthetic supervision for challenging transparent and fine-structure foregrounds. Experiments demonstrate state-of-the-art accuracy and boundary fidelity across standard benchmarks. We hope this framework and
dataset offer a scalable path toward high-fidelity matting in open-world scenes.

% We present RenderMatte, which successfully resolves the sparse-boundary bottleneck in image matting by reframing the task as a reward-aligned generative editing process. By combining an alpha-edge supervised flow-matching objective with GRPO-based reinforcement post-training, our model explicitly prioritizes fine-detail generation over visually safe but smoothed predictions. Furthermore, our proposed RenderMatte dataset provides the exact, strand-level supervision required to unleash this generative potential. Extensive experiments confirm that RenderMatte achieves state-of-the-art boundary fidelity and overall matting precision. We hope this framework and dataset pave the way for next-generation, structure-aware image matting.

% \begin{figure}[t]
%     \centering
%     \includegraphics[width=\columnwidth]{Figures/bad_cases.pdf}
%     \caption{
%     Failure cases on out-of-domain data. (Top to bottom: Input, Ours, GT). The model may produce suboptimal results when encountering rare foreground objects or extreme structures that deviate from the training distribution.
%     }
%     \label{fig:bad_cases}
% \end{figure}

\clearpage

\bibliography{aaai2027}

% % Check whether the conference requires a reproducibility checklist to be included in the paper.
% % If so, you can uncomment the following line and ajust the path to include it.
% % \input{ReproducibilityChecklist.tex}

\clearpage
\appendix
\begin{center}
    {\Large\bfseries Appendix}
\end{center}
\vspace{0.5em}

\begin{table*}[t]
\centering
\caption{Dataset statistics of RenderMatte.}
\label{tab:supp_dataset_statistics}
\setlength{\tabcolsep}{4pt}
\renewcommand{\arraystretch}{0.92}

\textbf{Foreground asset counts by source and category.}
\vspace{0.25em}

\resizebox{\textwidth}{!}{
\begin{tabular}{lrrrrrrrrr}
\toprule
Source & Person & Animal & Flower/Grass & Tree & Toy & Furniture & Transparent & Other & Total \\
\midrule
GPT-generated & 275 & 663 & 400 & 0 & 225 & 175 & 0 & 925 & 2,663 \\
Internet-collected & 3,479 & 4,459 & 3,002 & 737 & 61 & 5,701 & 457 & 0 & 17,896 \\
3D-rendered & 476 & 243 & 1,607 & 1,045 & 296 & 985 & 0 & 0 & 4,652 \\
\midrule
Total & 4,230 & 5,365 & 5,009 & 1,782 & 582 & 6,861 & 457 & 925 & 25,211 \\
\bottomrule
\end{tabular}
}

\vspace{0.8em}
\textbf{Composited training samples after pairing foreground assets with backgrounds.}
\vspace{0.25em}

\resizebox{\textwidth}{!}{
\begin{tabular}{lrrrrrrrrr}
\toprule
Source & Person & Animal & Flower/Grass & Tree & Toy & Furniture & Transparent & Other & Total \\
\midrule
GPT-generated & 2,200 & 5,304 & 3,200 & 0 & 1,800 & 1,400 & 0 & 7,400 & 21,304 \\
Internet-collected & 3,600 & 6,000 & 2,998 & 1,000 & 150 & 6,000 & 3,176 & 0 & 22,924 \\
3D-rendered & 6,440 & 8,000 & 6,340 & 9,570 & 6,780 & 2,175 & 0 & 0 & 39,305 \\
\midrule
Total & 12,240 & 19,304 & 12,538 & 10,570 & 8,730 & 9,575 & 3,176 & 7,400 & 83,533 \\
\bottomrule
\end{tabular}
}

\vspace{0.8em}
\textbf{Category distribution of the RenderMatte-2K benchmark.}
\vspace{0.25em}

\resizebox{\textwidth}{!}{
\begin{tabular}{lrrrrrrrrrrrrrrr}
\toprule
Category & People & Animal & Furniture & Toy & Tool & Sports & Accessory & Fruit & Musical & Textile & Craft & Transparent & Tree & Flower & Total \\
\midrule
Count & 220 & 220 & 170 & 150 & 140 & 140 & 130 & 130 & 130 & 130 & 120 & 110 & 110 & 100 & 2,000 \\
\bottomrule
\end{tabular}
}

\end{table*}

\paragraph{Overview.}

This appendix provides additional details that support the main paper. We first describe the construction of RenderMatte, including the foreground asset sources, foreground-background compositing statistics, and the category distribution of the RenderMatte-2K benchmark. We then report implementation details for supervised fine-tuning (SFT) and group-relative alpha alignment, including checkpoint selection, training settings, and reward-curve logging. Finally, we include representative failure cases to clarify the remaining limitations on out-of-domain examples.

\subsection{Dataset Construction}

RenderMatte is constructed from multi-source foreground assets and diverse
background compositions. Table~\ref{tab:supp_dataset_statistics} summarizes the
foreground asset sources, composited training samples, and the category
distribution of RenderMatte-2K. The foreground counts correspond to unique RGBA
assets from three sources: 3D-rendered assets, GPT-generated assets, and
Internet-collected assets. The composited counts are therefore image-mask pairs
rather than additional unique foreground categories.

The compositing ratio differs across sources. GPT-generated assets are expanded
by roughly eight backgrounds per asset. 3D-rendered assets are rendered and
composited more densely to increase viewpoint, lighting, pose, and boundary
diversity. Internet-collected assets are composited more selectively because the
source pool is already larger and more visually diverse. This design yields
25,211 foreground assets and 83,533 training composites.

We also construct RenderMatte-2K as a held-out benchmark with 2,000 samples from
unseen foregrounds and backgrounds. Its category distribution covers people,
animals, furniture, toys, tools, sports objects, accessories, fruits, musical
objects, textiles, crafts, transparent objects, trees, and flowers.

\subsection{Implementation Details}

\paragraph{Model architecture.}

RenderMatte uses FLUX.1 Kontext as an image-editing backbone. The text instruction is fixed across all samples, while the visual condition is provided by the original image and the trimap. The target image is the ground-truth alpha matte represented in an image-compatible format.

\paragraph{Supervised Fine-Tuning.}

The main paper reports the core supervised fine-tuning configuration. Here we provide additional implementation details that are not included in the main text.

\paragraph{Task formatting.}
Each training sample is formatted as a FLUX Kontext image-editing instance with a fixed instruction, ``Transform to matting map while maintaining original composition.'' The visual condition contains two reference images: the original RGB image and the trimap. The generation target is the ground-truth alpha matte. We use a fixed instruction rather than per-image captions, so the adaptation is driven by visual matting conditions rather than semantic text variation.

\paragraph{Online trimap construction.}
Trimaps are generated on the fly from the ground-truth alpha matte during training. We randomly sample the morphological kernel size from \([5,15)\) and the number of erosion/dilation iterations from \([5,15)\). The unknown region is defined as the pixels that remain outside both the eroded definite foreground and the dilated definite background. This online procedure exposes the model to different uncertain-band widths for the same underlying alpha matte.

\paragraph{Optimization schedule and checkpointing.}
The learning-rate schedule uses a short constant-scheduler warmup: the PyTorch \texttt{ConstantLR} default factor \(1/3\) is applied for the first five steps, after which the learning rate remains constant. Gradient checkpointing is enabled, and EMA is not used in the SFT stage. Checkpoints are saved every 2,000 steps. During training, a 10-sample AIM-500 quick evaluation is run every 500 steps, and the full 500-sample AIM-500 evaluation is run at each saved checkpoint.

\paragraph{SFT checkpoint selection.}
The SFT run saved full-parameter checkpoints at steps 2,000, 4,000, 6,000,
8,000, 10,000, and 12,000. Table~\ref{tab:supp_sft_checkpoint_results}
compares the checkpoints with complete benchmark metrics. The step-10,000
checkpoint is used as the initialization for the group-relative alpha alignment
stage. Although step 12,000 slightly improves AIM-500 MSE/MAD/SAD/Conn, it is
flat or mildly worse on P3M-500-NP and AM-2K, especially for boundary-sensitive
metrics. We therefore use step 10,000 as the more stable checkpoint for the
subsequent alignment stage.

\begin{table}[t]
    \centering
    \caption{SFT checkpoint comparison for checkpoints with complete benchmark metrics reported in the training record. Lower values are better for all metrics. Step 10,000 is used to initialize the alignment stage because it gives the more balanced validation behavior across benchmarks.}
    \label{tab:supp_sft_checkpoint_results}
    \resizebox{\columnwidth}{!}{
    \begin{tabular}{llccccc}
        \toprule
        Benchmark & Step & MSE & MAD & SAD & Grad & Conn \\
        \midrule
        P3M-500-NP & 10,000
        & \cellcolor{best}0.0010 & \cellcolor{best}0.0050 & \cellcolor{best}8.6069 & \cellcolor{best}8.6598 & \cellcolor{best}4.9296 \\
        P3M-500-NP & 12,000
        & 0.0011 & 0.0051 & 8.7953 & 9.3016 & 5.0567 \\
        \midrule
        AM-2K & 10,000
        & \cellcolor{best}0.0010 & \cellcolor{best}0.0056 & \cellcolor{best}9.6483 & \cellcolor{best}5.6652 & 5.7563 \\
        AM-2K & 12,000
        & \cellcolor{best}0.0010 & \cellcolor{best}0.0056 & 9.6894 & 5.8540 & \cellcolor{best}5.7362 \\
        \midrule
        AIM-500 & 10,000
        & 0.0025 & 0.0096 & 16.1219 & \cellcolor{best}11.3611 & 10.0452 \\
        AIM-500 & 12,000
        & \cellcolor{best}0.0023 & \cellcolor{best}0.0093 & \cellcolor{best}15.7202 & 11.5469 & \cellcolor{best}9.7252 \\
        \bottomrule
    \end{tabular}
    }
\end{table}

\subsection{Group-Relative Alpha Alignment}

The alignment stage starts from the SFT checkpoint at step 10,000. The base model is frozen, and only a rank-64 LoRA adapter on the DiT modules is trained. For each prompt-image condition, the sampler generates eight candidate mattes, forming the group used for relative advantage normalization. Training uses eight denoising steps, while evaluation uses one denoising step, both with guidance scale 1.0 and noise level 0.2. We do not use a KL penalty against the SFT reference policy in this stage.

The alignment run uses a constant learning rate without a scheduler. We enable activation checkpointing and per-prompt reward-statistic tracking. The SDE exploration is applied within a two-step window inside the eight-step sampling trajectory, which preserves stochastic exploration while limiting variance in long trajectories. Gradients are clipped with a maximum norm of 1.0, and EMA is maintained for the LoRA adapter with decay 0.9 and an update interval of 8 steps.

\subsection{Reward-Curve Logging}

\paragraph{Reward-Curve Protocol.}

The reward curve in the main paper is plotted from evaluation checkpoints under the same evaluation protocol, rather than from raw training summaries. In the inspected run directory, \texttt{reward\_history.jsonl} contains 43 training summaries and 21 evaluation summaries. Training summaries are recorded every 8 global steps over 128 sampled rollouts, while evaluation summaries are recorded every 16 global steps over 500 evaluation samples. LoRA checkpoints are saved every 16 steps from step 16 to step 320, and debug visualizations are stored for qualitative inspection. Table~\ref{tab:supp_reward_curve_points} reports the evaluation points used by the main reward curve through step 320.

\begin{table}[t]
    \centering
    \caption{Evaluation checkpoints available for the reward curve through step 320. Higher reward and lower error are better.}
    \label{tab:supp_reward_curve_points}
    \resizebox{\columnwidth}{!}{
    \begin{tabular}{lcccccc}
        \toprule
        Step & Reward & Error & MAD & SAD & Grad & Conn \\
        \midrule
        16 & -0.565 & 0.565 & 0.0167 & 4.371 & 3.721 & 2.241 \\
        32 & -0.577 & 0.577 & 0.0172 & 4.518 & 3.763 & 2.281 \\
        48 & -0.550 & 0.550 & 0.0164 & 4.304 & 3.700 & 2.232 \\
        64 & -0.558 & 0.558 & 0.0164 & 4.301 & 3.639 & 2.226 \\
        80 & -0.563 & 0.563 & 0.0167 & 4.385 & 3.731 & 2.249 \\
        96 & -0.555 & 0.555 & 0.0165 & 4.332 & 3.668 & 2.249 \\
        112 & -0.556 & 0.556 & 0.0163 & 4.272 & 3.715 & 2.213 \\
        128 & -0.568 & 0.568 & 0.0168 & 4.405 & 3.662 & 2.266 \\
        144 & -0.556 & 0.556 & 0.0165 & 4.322 & 3.679 & 2.234 \\
        160 & -0.547 & 0.547 & 0.0162 & 4.251 & 3.635 & 2.214 \\
        176 & -0.553 & 0.553 & 0.0163 & 4.265 & 3.642 & 2.256 \\
        192 & -0.546 & 0.546 & 0.0163 & 4.262 & 3.578 & 2.242 \\
        208 & -0.546 & 0.546 & 0.0160 & 4.191 & 3.605 & \cellcolor{best}2.211 \\
        224 & -0.548 & 0.548 & 0.0164 & 4.311 & 3.622 & 2.271 \\
        240 & -0.548 & 0.548 & 0.0163 & 4.281 & 3.612 & 2.244 \\
        256 & \cellcolor{best}-0.534 & \cellcolor{best}0.534 & \cellcolor{best}0.0158 & \cellcolor{best}4.150 & \cellcolor{best}3.550 & 2.225 \\
        272 & -0.542 & 0.542 & 0.0163 & 4.273 & 3.633 & 2.278 \\
        288 & -0.543 & 0.543 & 0.0163 & 4.277 & 3.621 & 2.253 \\
        304 & -0.544 & 0.544 & 0.0162 & 4.243 & 3.585 & 2.263 \\
        320 & -0.555 & 0.555 & 0.0166 & 4.339 & 3.660 & 2.267 \\
        \bottomrule
    \end{tabular}
    }
\end{table}

\subsection{Failure Cases}

Figure~\ref{fig:bad_cases} shows representative failure cases on out-of-domain data. These examples complement the limitation discussion in the main paper. Although RenderMatte benefits from exact-alpha synthetic supervision and reward-based alignment, it can still produce suboptimal mattes for rare foreground objects, unusual transparency patterns, or extreme structures that are underrepresented in the training distribution. Such cases suggest that broader foreground coverage and more diverse boundary/opacity patterns remain useful directions for future data construction.

\begin{figure*}[t]
    \centering
    \includegraphics[width=0.82\textwidth]{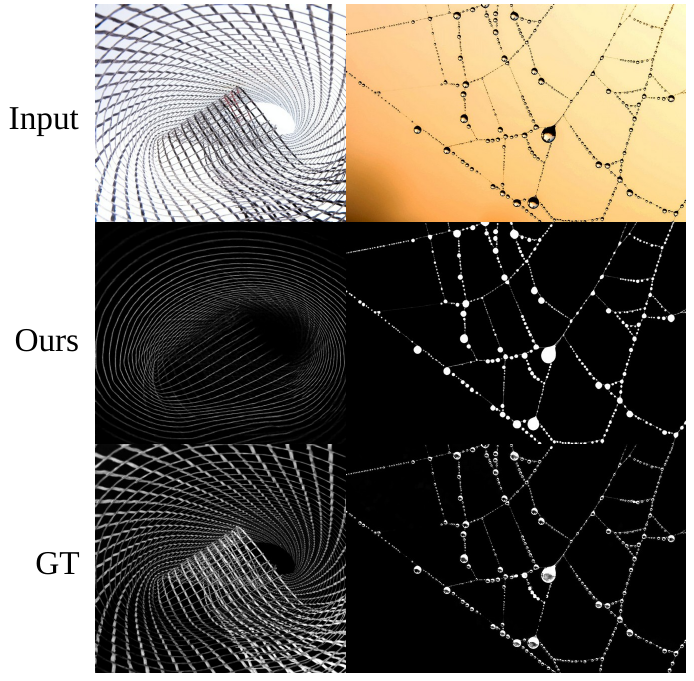}
    \caption{
    Failure cases on out-of-domain data. Top to bottom: input, RenderMatte, and
    ground truth. The model may produce suboptimal mattes for rare foreground
    objects or extreme structures that deviate from the training distribution.
    }
    \label{fig:bad_cases}
\end{figure*}

\end{document}